\documentclass[times,preprint,11pt,nopreprintline]{elsarticle}

\usepackage{amssymb}
\usepackage{amsmath}

\usepackage{booktabs}
\usepackage{makecell}
\usepackage{multirow}

\usepackage[margin=1in]{geometry}
\usepackage{array,float,caption}
\usepackage{setspace}

\usepackage[hidelinks]{hyperref}

\DeclareUrlCommand\path{\urlstyle{same}}
\biboptions{sort&compress}

\newdefinition{definition}{Definition}

\hypersetup{
  pdftitle={Learning Adaptive Deliberation Control for Reliable Recognition with Multi-Agent Language Models},
  pdfauthor={Yuan Liu; Haoyue Peng; Fanliang Hu; Lei Huang; Wenli Cai},
  pdfkeywords={Language-model recognition; Multi-agent control; False consensus; Ensemble diversity; Reinforcement learning}
}

\begin{document}

\begin{frontmatter}

\title{Learning Adaptive Deliberation Control for Reliable Recognition with Multi-Agent Language Models}

\author[cqu]{Yuan Liu}
\author[cqu]{Haoyue Peng}
\author[cqu]{Fanliang Hu}
\author[cqu]{Lei Huang}
\author[cqu]{Wenli Cai\corref{cor1}}
\ead{wenli.cai@cqu.edu.cn}

\affiliation[cqu]{organization={Medical Artificial Intelligence Laboratory, Chongqing University Cancer Hospital, Chongqing University},
                  city={Chongqing},
                  postcode={400030},
                  country={China}}

\cortext[cor1]{Corresponding author}

\begin{abstract}

Multi-agent deliberation can improve recognition by criticizing and revising candidate answers, but fixed controls cannot respond to changes in disagreement or reasoning quality. We present an adaptive multi-agent controller (ADMA) that learns continuous policies for proposal, challenge, arbitration, and coordination while keeping the task model frozen. Shared trace features describe quality, agreement, and progress, while role-specific features support decentralized control during inference. One controller trained on legal and financial reasoning is transferred without policy updates to four textual and multimodal benchmarks and two successive task-model versions. With Qwen3.5 held fixed, ADMA-Full improves accuracy by 6.2--14.2 percentage points over Direct and by 2.8--3.0 points over ADMA-Fixed. Paired bootstrap ranges for the Full--Fixed comparison are positive on three tasks. Outcome decompositions show that recoveries exceed introduced errors, while process assessment records fewer false-consensus events and more supported correct outcomes. Transfer to Qwen3.6 remains task dependent, which bounds the generality of the learned policy. Median token and latency ratios are 14.3--18.7 and 7.8--16.5 relative to Direct. These results show that adaptive deliberation can improve fixed-model recognition when its task-dependent gains justify the additional computation.

\end{abstract}


\end{frontmatter}

\section{Introduction}
\label{sec:introduction}

Language models increasingly address recognition tasks that require both interpretation and a discrete decision. Examples include identifying a legal holding, calculating an answer from a financial report, and selecting an option for a visual question. Multi-agent systems distribute proposal, critique, coordination, and adjudication across specialized roles \cite{tang2024medagents,sapkota2026agentic}. Existing systems vary their communication structures and workflows, but they share the aim of correcting errors through interaction \cite{wang2024survey}.

Collaboration is useful only when the participants contribute complementary information. Classifier fusion benefits from complementary predictions \cite{kittler1998combining}, whereas agreement among correlated predictors provides little independent confirmation \cite{kuncheva2003measures}. This distinction matters when language-model roles share a backbone and input context. Different prompts may produce similar rationales, allowing an error to survive repeated agreement. We call an incorrect, concentrated decision without adequate independent support \emph{semantic false consensus}. A deliberative system must therefore elicit informative alternatives and determine whether they strengthen the decision.

Additional reasoning can be organized through multi-step or uncertainty-aware inference \cite{plaat2025multistep,shorinwa2025uncertainty}, self-refinement \cite{madaan2023self}, or multi-agent debate \cite{tian2024encouraging}. These methods improve how candidate answers are generated and compared, but each case may require a different amount of criticism and revision. Goal-driven and collaborative frameworks define roles and workflows \cite{sado2023agents,anjos2023collaborative}, while model routing selects the models assigned to them \cite{yue2025masrouter}. Neither choice determines how a fixed group should adapt its behavior as disagreement, quality, and remaining computation change. This unresolved control problem motivates the present study.

We propose ADMA, an adaptive controller comprising four frozen roles: proposer, challenger, arbiter, and coordinator. Role-specific policies use trace quality, agreement, and progress to regulate each role. Centralized critics learn the effects of joint actions during training. At inference, each actor uses its own observation while the task model remains unchanged.

The evaluation follows four linked questions. We first test whether ADMA improves a fixed task model relative to Direct and fixed-control deliberation. We then examine whether one trained controller transfers across tasks, modalities, and a successive model version. Paired outcomes and process diagnostics identify where gains arise and whether correct decisions receive adequate support. Resource measurements finally determine the computational demand associated with those gains.

This study contributes a continuous control formulation for frozen language-model roles, conditional diagnostics for false consensus and supported correctness, and a matched evaluation of accuracy, correction, reliability, and cost. Together, these elements test adaptive deliberation as an inference policy rather than as a change to the underlying recognition model.

\section{Related work}
\label{sec:related}

\subsection{Adaptive cooperation in recognition systems}

Ensemble methods combine predictors with complementary errors \cite{dietterich2000ensemble}, while dynamic selection adapts the contributing classifiers to each input \cite{cruz2018dynamic}. DES-AS, for example, estimates cooperative contributions for classifier selection and weighting \cite{zhang2025desas}. Selective recognition similarly conditions prediction or deferral on expected error \cite{chow1970optimum,hasan2025reject}. ADMA extends case-dependent selection from classifier outputs to the behavior of interacting language-model roles.

The controller also requires signals that distinguish useful revision from repetition. Confidence and uncertainty are central to machine-learning safety \cite{mohseni2022safety}, uncertainty-aware ensembles, and out-of-distribution detection \cite{arco2023ensembles,lu2025ood}. Deliberation adds temporal information because the value of another round depends on how arguments have changed. ADMA therefore combines confidence with support, agreement, quality changes, and remaining rounds.

\subsection{Language-model recognition and multi-agent reasoning}

Language-model recognition spans distinct reasoning demands. CaseHOLD and LexGLUE test legal interpretation \cite{zheng2021does,chalkidis2022lexglue}, whereas FinQA requires calculations grounded in financial reports \cite{chen2021finqa}. Chain-of-thought and prompting methods expose intermediate reasoning \cite{wei2022chain,liu2023prompting}. Pretrained-language-model pipelines also incorporate external actions \cite{min2023pretrained}, and tool-augmented models can learn when to invoke functions \cite{schick2023toolformer}. These methods construct individual reasoning paths; ADMA instead controls how several paths are proposed, challenged, and adjudicated.

Other methods improve the information available to recognition. PLRTQA combines retrieval-augmented generation with transfer learning for textbook question answering \cite{alawwad2025textbook}. Xia et al. combine reasoning elicitation with multi-granularity contrastive learning for text-rich images \cite{xia2026reasoning}. These methods enrich context or visual--semantic alignment, whereas ADMA addresses how multiple roles examine the available information.

Debate subjects claims to criticism, while round-table reconciliation combines competing responses before a final decision \cite{wang2023can,chen2024reconcile}. Sparse communication changes how disagreement is exchanged \cite{li2024improving}. ADMA keeps the communication structure fixed and adapts the intensity of proposal, challenge, and adjudication from the evolving trace.

MoXpert uses adaptive gating to coordinate retrieval, domain knowledge, reasoning, and decision modules for industrial anomaly detection \cite{chen2026moxpert}. ADMA learns continuous role control through centralized training and decentralized execution, a standard multi-agent reinforcement-learning formulation \cite{geng2026marl}. Critics evaluate joint interaction during training, while each actor uses its own observation at inference. This separation learns interdependent behavior without changing the task model or role structure.

\subsection{Reliability assessment for generated reasoning}

Final-answer accuracy does not establish whether a prediction is well supported. Predictive-uncertainty explanations identify feature and interaction effects \cite{bley2025uncertainty}. SelfCheckGPT and semantic entropy use response variation to assess hallucination or uncertainty \cite{manakul2023selfcheckgpt,farquhar2024detecting}, while G-Eval applies explicit criteria to generated content \cite{liu2023geval}. We preserve this distinction by deriving correctness from benchmark answers and assessing the reasoning trace separately. False consensus characterizes unsupported errors, whereas evidence support characterizes adequately justified correct predictions.

\section{Problem formulation}
\label{sec:problem}

Given a textual or multimodal input $x$, the recognition system produces a prediction $\hat y\in\mathcal{Y}$, or a task-normalized numerical answer, and an interaction trace $H_T$. The proposer constructs a candidate answer, the challenger examines its weaknesses and alternatives, the arbiter evaluates the competing arguments, and the coordinator regulates the interaction. All four roles use frozen instances of the task model. The learning problem is to determine their behavior from the information accumulated during deliberation.

We model deliberation as a finite-horizon decision process. At round $t$, state $s_t$ summarizes the available trace, and role $i$ receives observation $o_t^i$. Its policy produces action $a_t^i=\mu_i(o_t^i)$, which modifies the role instructions and interaction conditions. The resulting responses update the trace and the next state. Deliberation ends under the specified termination rule or at horizon $T_{\max}$. This formulation couples the next interaction to the progress made so far.

\begin{definition}[Semantic false consensus]
For an evaluated question, semantic false consensus combines an incorrect final prediction with concentrated commitment and inadequate independent support under the prespecified assessment. The false-consensus rate (FCR) is the proportion of incorrect predictions that meet these conditions. The evidence-support rate (ESR) is the proportion of correct predictions judged to have adequate support. Both metrics are conditional on the final outcome.
\end{definition}

Benchmark answers determine correctness; FCR and ESR assess the associated reasoning and are reserved for evaluation. Section~\ref{sec:outcomes} specifies their operational criteria.

\section{Adaptive multi-agent controller}
\label{sec:method}

\subsection{Overview}

\begin{figure}[!t]
\centering
\includegraphics[width=\textwidth]{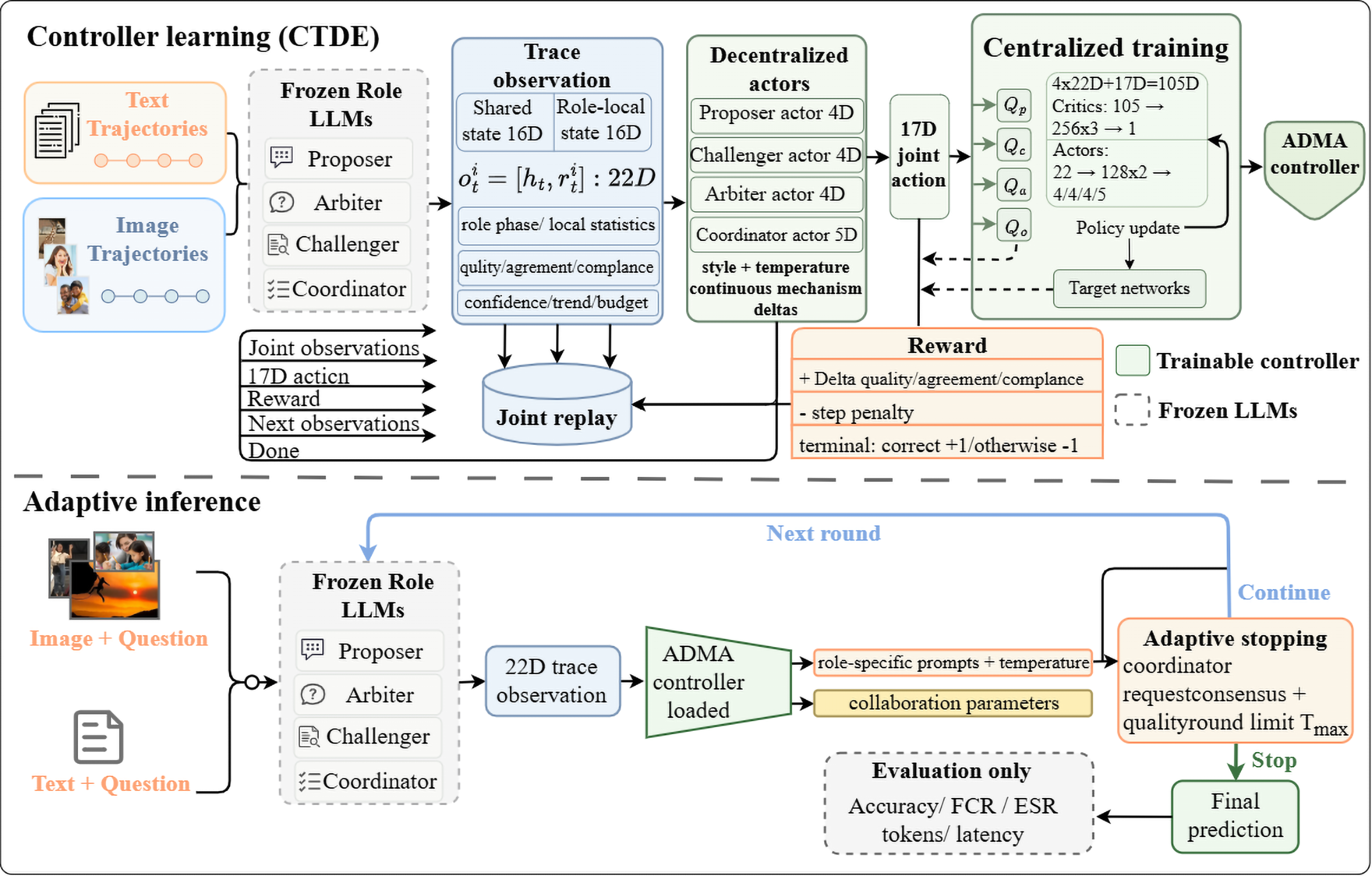}
\caption{ADMA controller learning and adaptive inference. During centralized training, joint transitions from four frozen role models are stored in replay and evaluated by role-specific critics. During decentralized inference, the learned actors adapt role behavior and collaboration from the evolving trace until termination. The final prediction is evaluated separately for accuracy, reliability, and resource use.}
\label{fig:architecture}
\end{figure}

Figure~\ref{fig:architecture} summarizes the controller. Shared and role-specific trace observations determine continuous actions that condition the next response round. Centralized critics learn from joint transitions, while role-specific actors regulate the frozen task model during inference.

\subsection{Trace observation}

The controller must distinguish productive disagreement from repetition and premature convergence. Its observation therefore represents both the current assessment and its change over time. Given trace $H_t$, a deterministic mapping forms shared summary $h_t=f(H_t)$, and role $i$ receives
\begin{equation}
\label{eq:trace_observation}
o_t^i=[h_t,r_t^i],
\end{equation}
where $h_t$ contains 16 shared features and $r_t^i$ contains six role-specific features, yielding a 22-dimensional observation. Shared features describe progress, remaining rounds, quality, agreement, compliance, support, interaction mode, intensity, and proposer--challenger assessments. Role-specific features describe phase, assertiveness, quality contribution, compliance, recent confidence, and activity. Progress is normalized by the six-round horizon, while changes in quality and agreement are clipped to $[-1,1]$. The two components provide a common trace summary and the information needed for role-specific control.

\subsection{Continuous role control}

Continuous actions adjust each role's behavior without changing its function. The proposer, challenger, and arbiter each produce four actions; the coordinator produces five. Actor outputs $a\in[-1,1]$ are mapped to $u=(a+1)/2$ and translated into role-specific instructions and decoding temperatures.

Proposer controls regulate assertiveness, detail, rule adherence, and feedback use. Challenger controls regulate intensity, specificity, constructiveness, and novelty. Arbiter controls regulate strictness, feedback detail, consensus preference, and rule adherence. The coordinator adjusts interaction intensity, consensus sensitivity, mode-switching thresholds, and exploration. Decoding temperatures are bounded to 0.3--1.2 for proposal and challenge and 0.2--0.8 for arbitration. The same role structure can therefore produce different deliberation patterns as the trace changes.

\subsection{Policy learning}

The learning objective rewards decision improvement while penalizing unproductive continuation. It combines trace development with terminal correctness. The shared shaping term is $b_t=0.3\Delta q_t+0.2\Delta g_t-0.01$, where $q_t$ and $g_t$ denote assessed quality and agreement. Role-specific terms reward valid proposals, informative challenges, coherent adjudication, and compliant interaction. Challenger reward increases when disagreement improves quality and decreases when disagreement lowers quality. Additional terms capture novelty, convergence, and diminishing returns. Terminal benchmark correctness contributes $+1$ for a correct prediction and $-1$ otherwise. FCR, ESR, token use, and latency remain evaluation outcomes rather than training rewards.

The actions of one role affect the observations and rewards of the others. We account for this dependence through MADDPG with centralized training and decentralized execution. Actor $\mu_i(o^i)$ uses the observation of role $i$, while critic $Q_i(o^1,\ldots,o^4,a^1,\ldots,a^4)$ evaluates the joint observations and actions. For a replay transition, the critic target is
\begin{equation}
\label{eq:critic_target}
y_i=r_i+\gamma(1-d)Q_i'\!\left(o',\mu_1'(o^{1'}),\ldots,\mu_4'(o^{4'})\right).
\end{equation}
In Eq.~\ref{eq:critic_target}, $\gamma$ is the discount factor, $d$ indicates episode termination, and primes denote the next observation or target networks, as appropriate. The actor maximizes its critic value with the other roles' replay actions held fixed. Polyak averaging updates the target networks.

\subsection{Adaptive inference}

At inference, trained actors condition the frozen roles on current observations and update their actions after each response. ADMA-Full requires challenger-guided revision and at least two rounds in the evaluated configuration. After a valid decision, deliberation may terminate through the coordinator. It may also stop when quality reaches 0.80 and the trace shows explicit consensus or disagreement below 0.15. A six-round horizon bounds the process. These rules preserve an opportunity to revise the initial proposal while limiting continued interaction.

The main computational demand comes from repeated task-model calls. If $N_t$ roles are active at round $t$ and $L_t^i$ is the context length of role $i$, the principal cost is $\sum_t\sum_{i=1}^{N_t}C_{\mathrm{LLM}}(L_t^i)$. Role participation, accumulated context, and deliberation depth therefore determine most of the additional inference cost. We measure token use and elapsed time alongside recognition performance to characterize this trade-off.

\begin{figure}[!t]
\centering
\includegraphics[width=\textwidth]{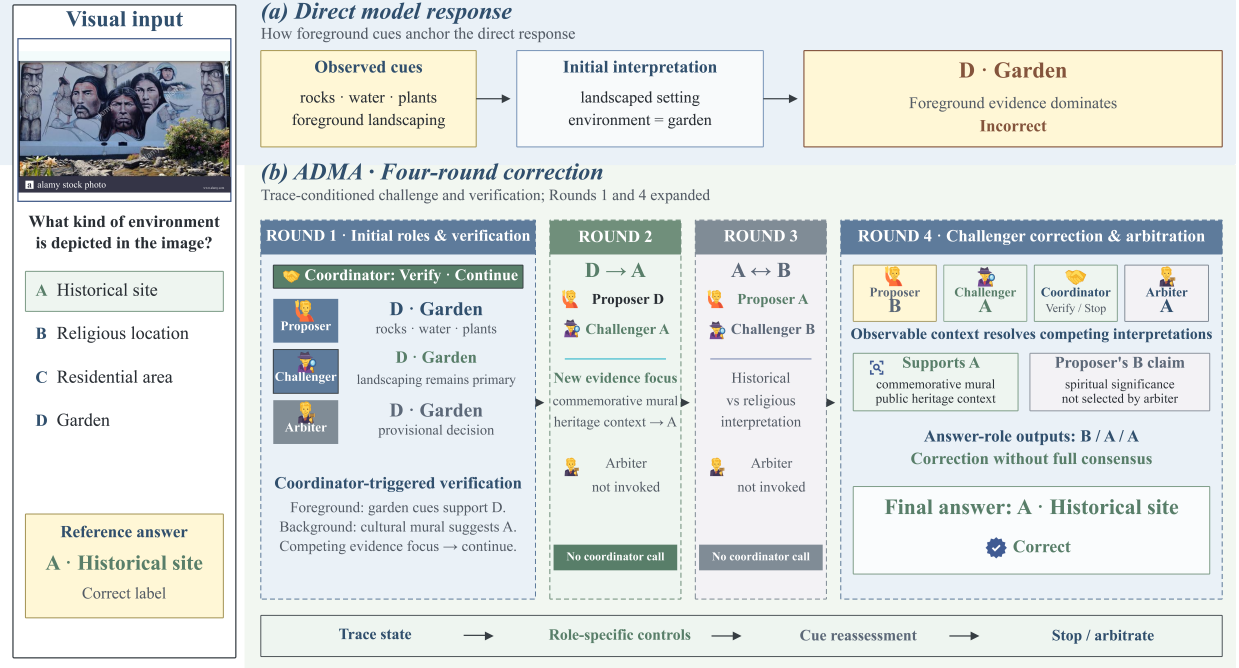}
\caption{Correction of MMStar item 33 in an archived Gemini 3.1 run. \textbf{a}, Direct selects D from foreground landscaping cues. \textbf{b}, the coordinator redirects verification toward the mural and historical context; four rounds of challenge and arbitration yield A while the final answering roles remain divided as B/A/A. This qualitative trace is separate from the aggregate Qwen3 evaluation.}
\label{fig:qualitative_correction}
\end{figure}

The example in Fig.~\ref{fig:qualitative_correction} illustrates how deliberation changes the basis of a decision. Direct inference classifies the scene as a garden by emphasizing foreground rocks, water, and vegetation. The coordinator directs attention to the mural and historical setting, and challenge and revision introduce the historical-site alternative. Subsequent rounds compare this alternative with the remaining religious interpretation. Final arbitration selects the correct option while the answering roles retain different judgments. The trace illustrates the distinction between directing attention to relevant cues and obtaining unanimous agreement.

\section{Experimental protocol}
\label{sec:experiments}

The experiments follow the four questions posed in the Introduction. They test within-model improvement, controller transfer, the paired structure and reliability of changed outcomes, and the associated computational demand. All protocols use matched questions, and every ADMA-Full evaluation uses the same trained controller.

\subsection{Datasets and evaluation design}

We evaluate legal, financial, and multimodal recognition using four public benchmarks. CaseHOLD, accessed through LexGLUE and denoted LawQA, assesses legal holding prediction. FinQA requires numerical reasoning over financial reports. MMBench \cite{liu2024mmbench} and MMStar \cite{chen2024mmstar} assess multimodal multiple-choice reasoning. Each benchmark contributes a fixed panel of 250 questions shared across the applicable model and protocol comparisons.

MMBench uses strict CircularEval, under which a base question is correct only if every available option permutation is answered correctly. The other benchmarks use item-level correctness after task-specific answer normalization. Because MMStar includes items sourced from MMBench, the two visual panels are not treated as independent samples.

Each model--protocol combination is evaluated twice on the same questions, giving 500 aligned question--repeat observations per task. We report the mean and sample standard deviation of the two run-level results to describe repeat variability. The base question remains the unit for paired resampling.

\subsection{Task models and comparison protocols}

The task models are Qwen3.5-35B-A3B and Qwen3.6-35B-A3B, successive versions from the same series and size class. Direct produces one response for each ordinary item or CircularEval permutation. ADMA-Fixed uses four roles with constant controls. ADMA-Full applies learned controls and challenger-guided revision under a six-round horizon. ADMA-Fixed is evaluated with Qwen3.5 and may terminate after one round. The Full--Fixed comparison therefore evaluates the complete adaptive protocol, including its revision requirement, rather than isolating one control component. Token and latency measurements quantify each protocol's resource demand.

Both model versions were evaluated under a common inference environment. Direct inference uses fixed low-temperature decoding, whereas ADMA adapts generation temperature across roles and rounds. Maximum response lengths are matched across versions on text tasks and are version-specific on visual tasks, so cross-version results describe the evaluated model--protocol combinations. Process assessment uses Gemini-3.1-Flash-Lite-Preview for Qwen3.5 and Gemini-3.1-Flash-Lite for Qwen3.6. FCR and ESR comparisons are made within each task-model version under its corresponding assessor; Qwen3.6 LawQA assessments are unavailable.

\subsection{Controller training and transfer}

Controller training uses 500 episodes drawn from 250 LawQA and 250 FinQA training questions. The development pool contains 50 questions per task. A fixed, balanced subset of 20 evaluates checkpoints every 100 episodes. Validation accuracy selects the checkpoint, with final quality resolving ties, yielding episode 400. Training, development, and test panels have disjoint identifiers within each textual benchmark. MADDPG uses a discount factor of 0.95, a Polyak coefficient of 0.01, and minibatches of 256 transitions.

The training roles use frozen Gemini-3.1-Flash-Lite-Preview models. To examine transfer, we apply the selected controller to both Qwen versions without further policy updates. The visual benchmarks are used only for evaluation. This design tests whether a policy learned from textual interactions can regulate different task models and multimodal reasoning. Both inference repeats use the same selected controller.

\subsection{Outcomes and statistical analysis}
\label{sec:outcomes}

Accuracy follows the evaluation unit of each benchmark. Inference failures count as incorrect within the fixed denominator, including CircularEval questions with a failed required permutation. FCR and ESR use the incorrect and correct subsets, respectively. Their changes describe reasoning quality conditional on the final outcome.

For FCR, concentrated commitment is recorded when mean final-answer confidence reaches 0.80, semantic answers are identical across CircularEval permutations, or the trace explicitly reports consensus. Under single-response Direct inference, the criterion identifies an unsupported error accompanied by high confidence or consistency across permutations. It provides the single-response analogue of the multi-role diagnostic.

The assessor receives the input, answer options, method-neutral prediction trace, and image when applicable, without the reference answer. Adequate support requires relevant facts, legal reasoning, visual observations, or verifiable calculations. Independent support requires at least two substantively different and independently checkable reasoning paths; paraphrases and repeated premises do not qualify. Only completed predictions with completed assessments can register either event, and unavailable assessments remain missing.

Accuracy and ESR differences are computed as ADMA-Full minus the comparator, while FCR reduction uses the opposite direction so that positive changes favor ADMA-Full. For paired bootstrap analysis, we average each question's effect across the two repeats and resample the 250 question clusters 10,000 times. The 2.5th and 97.5th percentiles describe sensitivity to the composition of the evaluated panel. Descriptive subgroup analyses partition FinQA by operation, information source, and reasoning depth and the visual tasks by benchmark category.

Resource analysis pairs observations for the same questions. For CircularEval, token use and elapsed inference time are summed across permutations, and deliberation depth is the maximum round count. After excluding failure-affected pairs, per-question ratios pool both repeats and are summarized by the median, interquartile range, and 10th--90th percentile span. We also divide the total resource increment over the complete panel by the net additional correct count. This measure includes expenditure on every outcome and quantifies the cost associated with an aggregate recognition gain.

\section{Results}
\label{sec:results}

\subsection{Within-model recognition performance}

We first ask whether adaptive deliberation improves recognition when the task model is fixed. Table~\ref{tab:raw_strength_results} shows that ADMA-Full achieves the highest Qwen3.5 accuracy on all four tasks. It gains 6.2--14.2 percentage points over Direct and 2.8--3.0 points over ADMA-Fixed. Because Full and Fixed use the same roles, their difference measures the complete adaptive protocol's added value under the stated scheduling difference.

The Full--Fixed mean difference is positive on every task. Its 95\% paired bootstrap range is positive on LawQA [0.4, 5.8], FinQA [0.2, 6.0], and MMBench [0.8, 5.4]. The MMStar range [$-1.4$, 7.0] includes zero. Adaptive control therefore improves the observed mean consistently, although the MMStar estimate remains sensitive to panel composition.

The same controller transfers unevenly to Qwen3.6. ADMA-Full gains 14.4 points on FinQA, 2.2 on MMBench, and 9.4 on MMStar, but loses 2.4 on LawQA. Bootstrap ranges are positive for FinQA and MMStar and include zero for LawQA and MMBench. Transfer therefore preserves substantial gains on selected tasks rather than establishing a model-independent benefit.

\begin{table}[!htbp]
\centering
\caption{Recognition performance across model versions and inference protocols.}
\label{tab:raw_strength_results}
\rmfamily\normalsize
\setlength{\tabcolsep}{2pt}
\begin{tabular*}{\textwidth}{@{\extracolsep{\fill}}lllccc@{}}
\toprule
Task model & Dataset & Protocol & \makecell{Accuracy\\(\%)} & \makecell{FCR\\(\%)} & \makecell{ESR\\(\%)} \\
\midrule
Qwen3.6 & LawQA & Direct & 77.2 $\pm$ 0.6 & -- & -- \\
 & LawQA & ADMA-Full & 74.8 $\pm$ 1.1 & -- & -- \\
 & FinQA & Direct & 57.4 $\pm$ 1.4 & 80.1 $\pm$ 11.3 & 96.2 $\pm$ 0.6 \\
 & FinQA & ADMA-Full & 71.8 $\pm$ 0.8 & 63.9 $\pm$ 5.9 & 98.3 $\pm$ 0.8 \\
 & MMBench & Direct & 74.0 $\pm$ 1.1 & 83.9 $\pm$ 2.6 & 95.7 $\pm$ 0.7 \\
 & MMBench & ADMA-Full & 76.2 $\pm$ 0.3 & 42.9 $\pm$ 8.8 & 95.5 $\pm$ 0.4 \\
 & MMStar & Direct & 49.4 $\pm$ 3.1 & 92.9 $\pm$ 0.7 & 90.6 $\pm$ 2.3 \\
 & MMStar & ADMA-Full & 58.8 $\pm$ 1.1 & 54.9 $\pm$ 2.2 & 87.8 $\pm$ 1.7 \\
\midrule
Qwen3.5 & LawQA & Direct & 59.0 $\pm$ 1.4 & 96.6 $\pm$ 0.6 & 44.4 $\pm$ 1.3 \\
 & LawQA & ADMA-Fixed & 70.2 $\pm$ 1.4 & 86.7 $\pm$ 3.2 & 54.1 $\pm$ 1.3 \\
 & LawQA & ADMA-Full & 73.2 $\pm$ 0.6 & 76.9 $\pm$ 4.8 & 61.5 $\pm$ 1.5 \\
 & FinQA & Direct & 57.0 $\pm$ 1.4 & 95.3 $\pm$ 1.5 & 63.8 $\pm$ 1.4 \\
 & FinQA & ADMA-Fixed & 67.8 $\pm$ 0.3 & 63.4 $\pm$ 2.3 & 78.8 $\pm$ 0.9 \\
 & FinQA & ADMA-Full & 70.8 $\pm$ 0.6 & 47.3 $\pm$ 0.1 & 82.8 $\pm$ 0.5 \\
 & MMBench & Direct & 65.6 $\pm$ 1.1 & 90.1 $\pm$ 1.1 & 66.5 $\pm$ 0.3 \\
 & MMBench & ADMA-Fixed & 71.0 $\pm$ 0.3 & 60.0 $\pm$ 2.3 & 79.4 $\pm$ 1.1 \\
 & MMBench & ADMA-Full & 74.0 $\pm$ 0.6 & 56.1 $\pm$ 2.0 & 85.1 $\pm$ 1.3 \\
 & MMStar & Direct & 47.4 $\pm$ 0.8 & 96.9 $\pm$ 1.1 & 54.0 $\pm$ 1.4 \\
 & MMStar & ADMA-Fixed & 50.8 $\pm$ 0.0 & 68.3 $\pm$ 2.3 & 58.7 $\pm$ 1.7 \\
 & MMStar & ADMA-Full & 53.6 $\pm$ 1.1 & 53.9 $\pm$ 0.7 & 67.9 $\pm$ 1.7 \\
\bottomrule
\end{tabular*}
\par\smallskip
{\footnotesize\noindent Values are mean $\pm$ sample standard deviation across two runs on the same 250-question panel; the 500 observations are not independent. FCR is the false-consensus rate among incorrect predictions, and ESR is the evidence-support rate among correct predictions; -- denotes an unavailable assessment. Qwen3.5 and Qwen3.6 denote Qwen3.5-35B-A3B and Qwen3.6-35B-A3B, respectively.\par}
\end{table}

\subsection{Cross-version recognition gains}

\begin{figure}[!t]
\centering
\includegraphics[width=\textwidth]{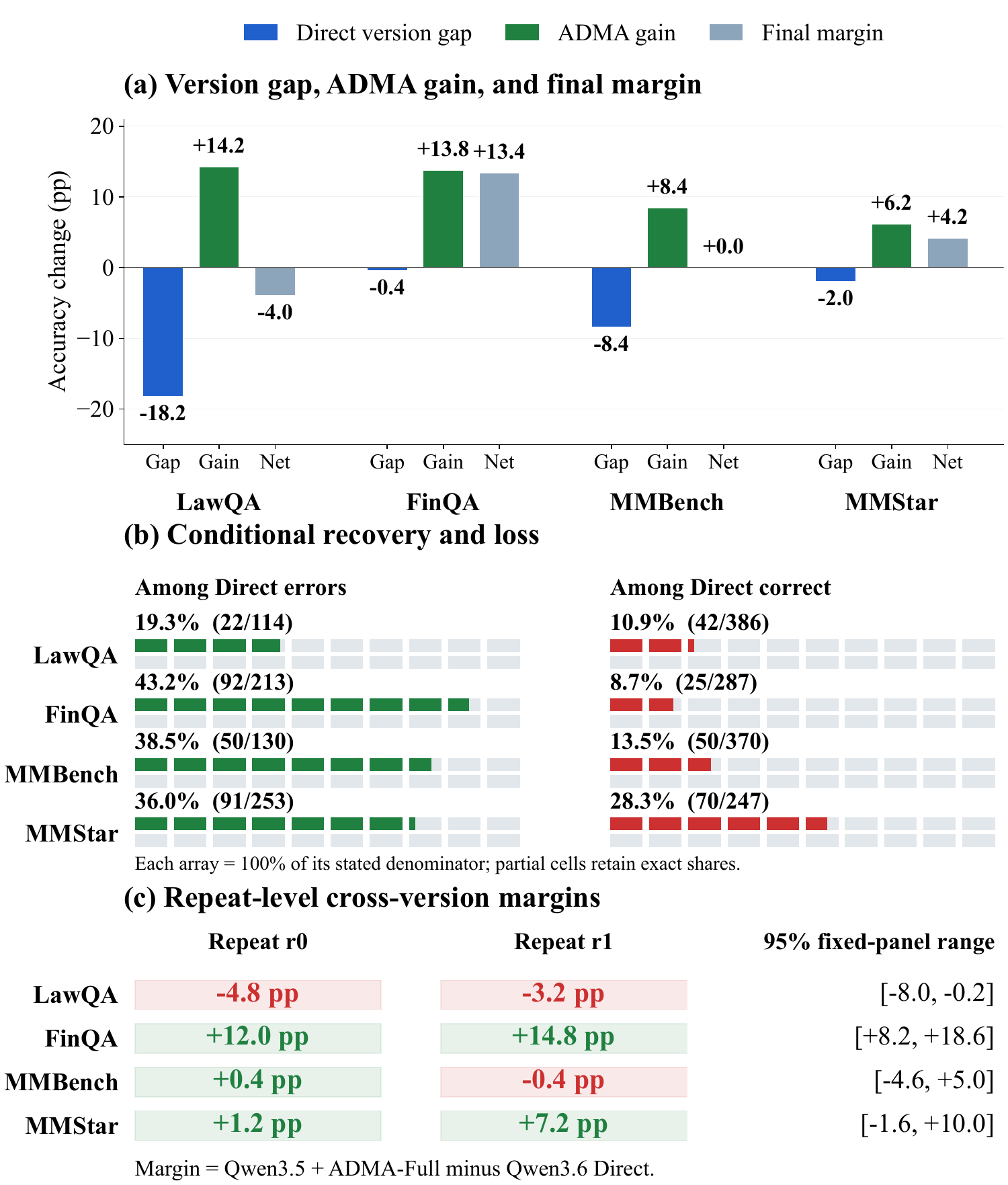}
\caption{Interaction gains relative to a successive model-version reference. \textbf{a}, the signed Direct version gap, within-Qwen3.5 ADMA gain, and resulting margin over Qwen3.6 Direct are shown from a common zero baseline. \textbf{b}, paired recovery among Qwen3.6 Direct errors and paired loss among its correct predictions, with counts and conditional denominators. \textbf{c}, repeat-specific final margins and 95\% question-cluster bootstrap ranges for the fixed-panel means. Qwen3.5 and Qwen3.6 denote Qwen3.5-35B-A3B and Qwen3.6-35B-A3B.}
\label{fig:raw_accuracy}
\end{figure}

We next ask whether adaptive deliberation can offset a successive-version Direct gap. Figure~\ref{fig:raw_accuracy}(a) compares the Qwen3.5 interaction gain with that gap. Qwen3.5 ADMA-Full exceeds Qwen3.6 Direct by 13.4 points on FinQA and 4.2 on MMStar. It matches MMBench and remains 4.0 points lower on LawQA. These task-specific margins show that interaction can sometimes outweigh a version change, but does not uniformly replace it.

Matched predictions explain the aggregate margins. On FinQA, Qwen3.5 ADMA-Full recovers 92 of 213 Qwen3.6 Direct errors while losing 25 of 287 correct observations. The balance produces 67 additional correct outcomes. MMBench instead contains 50 method-exclusive correct outcomes on each side despite equal mean accuracy. Cross-version parity can therefore conceal complementary error patterns.

FinQA remains positive in both repeats and throughout its bootstrap range. MMStar is positive in both repeats, but its range includes zero. MMBench varies around parity, whereas LawQA remains negative. FinQA thus provides the strongest evidence that deliberation can offset the evaluated version gap.

\subsection{Paired outcome structure and task heterogeneity}

\begin{figure}[!t]
\centering
\includegraphics[width=\textwidth]{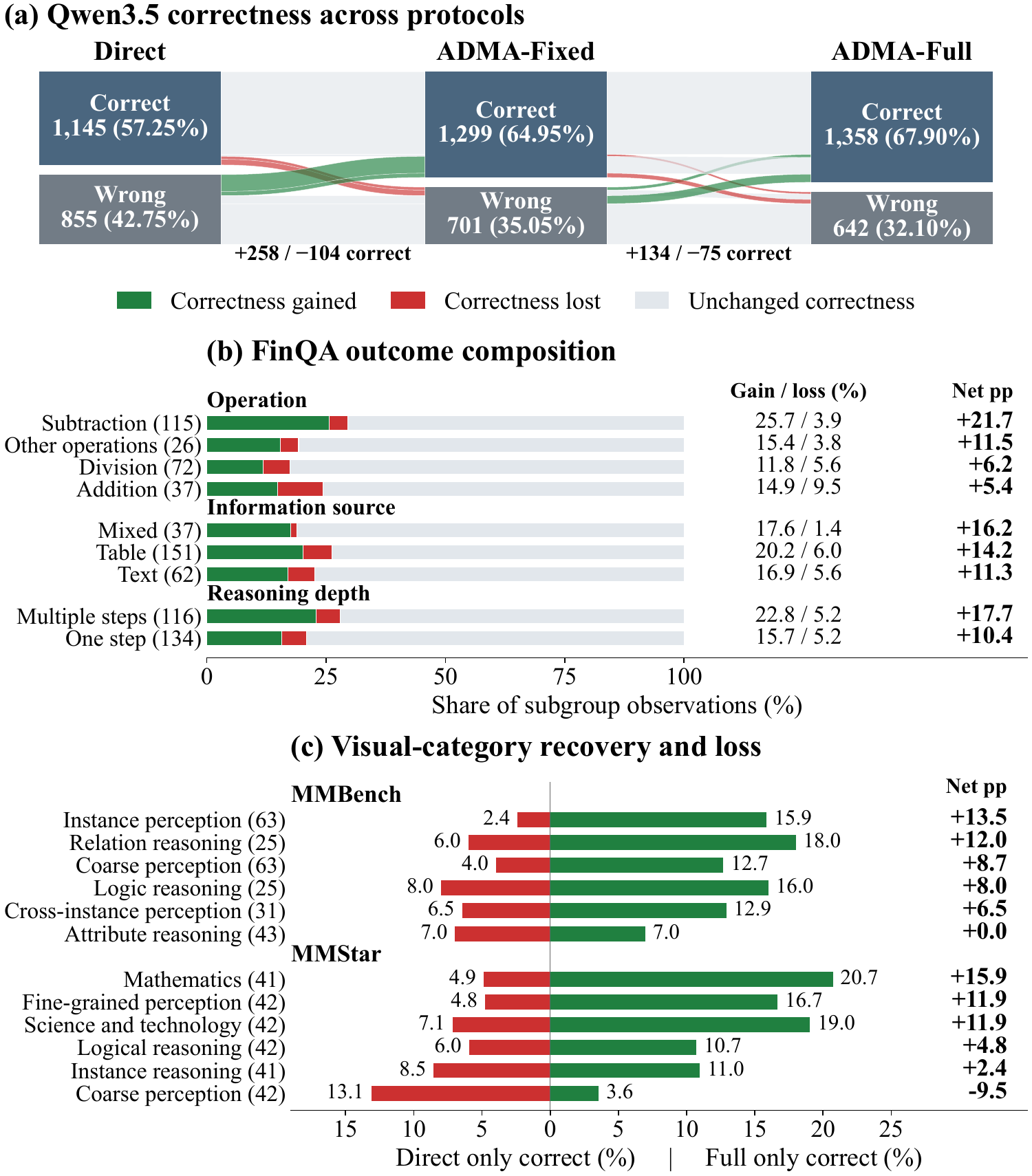}
\caption{Paired correctness composition within Qwen3.5. \textbf{a}, eight joint patterns across Direct, ADMA-Fixed, and ADMA-Full for 2,000 task--question--repeat observations. \textbf{b}, ADMA-Full versus Direct on FinQA, partitioned separately by operation, information source, and reasoning depth. \textbf{c}, the same comparison by visual benchmark category. Green, red, and gray denote gained, lost, and unchanged correctness. Subgroup percentages pool both repeats; parentheses give base-question counts.}
\label{fig:accuracy_differences}
\end{figure}

We then ask which matched outcomes produce the within-model gains. Figure~\ref{fig:accuracy_differences}(a) summarizes 2,000 Qwen3.5 task--question--repeat observations. Relative to Fixed, Full recovers 134 outcomes and loses 75, yielding 59 additional correct predictions. The flows compare complete protocol runs on matched questions; they are not successive stages of one trace.

FinQA gains vary with reasoning demand. Multi-step questions improve by 17.7 points, compared with 10.4 points for one-step questions. Their loss shares are similar, but multi-step questions have more recoveries. Subtraction and mixed-source questions also gain more than addition, summation, or text-only questions. These patterns associate deliberation with tasks that expose intermediate operations or multiple information sources for checking.

Five of six categories improve within each visual benchmark. MMBench attribute reasoning has balanced recovery and loss, while MMStar coarse perception declines. On MMStar, mathematics gains 15.9 points and coarse perception loses 9.5. Benchmark averages therefore conceal task categories in which deliberation helps or harms recognition.

\subsection{Reliability and deliberation allocation}

We next ask whether changed predictions are accompanied by changed reasoning support. Under the common Qwen3.5 assessment, ADMA-Full has fewer false-consensus events and more supported correct outcomes than Direct on every task. On FinQA, 148 observations exhibit false consensus only under Direct, compared with 12 only under Full. Supported correct outcomes increase by a net 111. The accuracy gain therefore coincides with fewer unsupported errors and more adequately supported correct decisions.

\begin{figure}[!t]
\centering
\includegraphics[width=\textwidth]{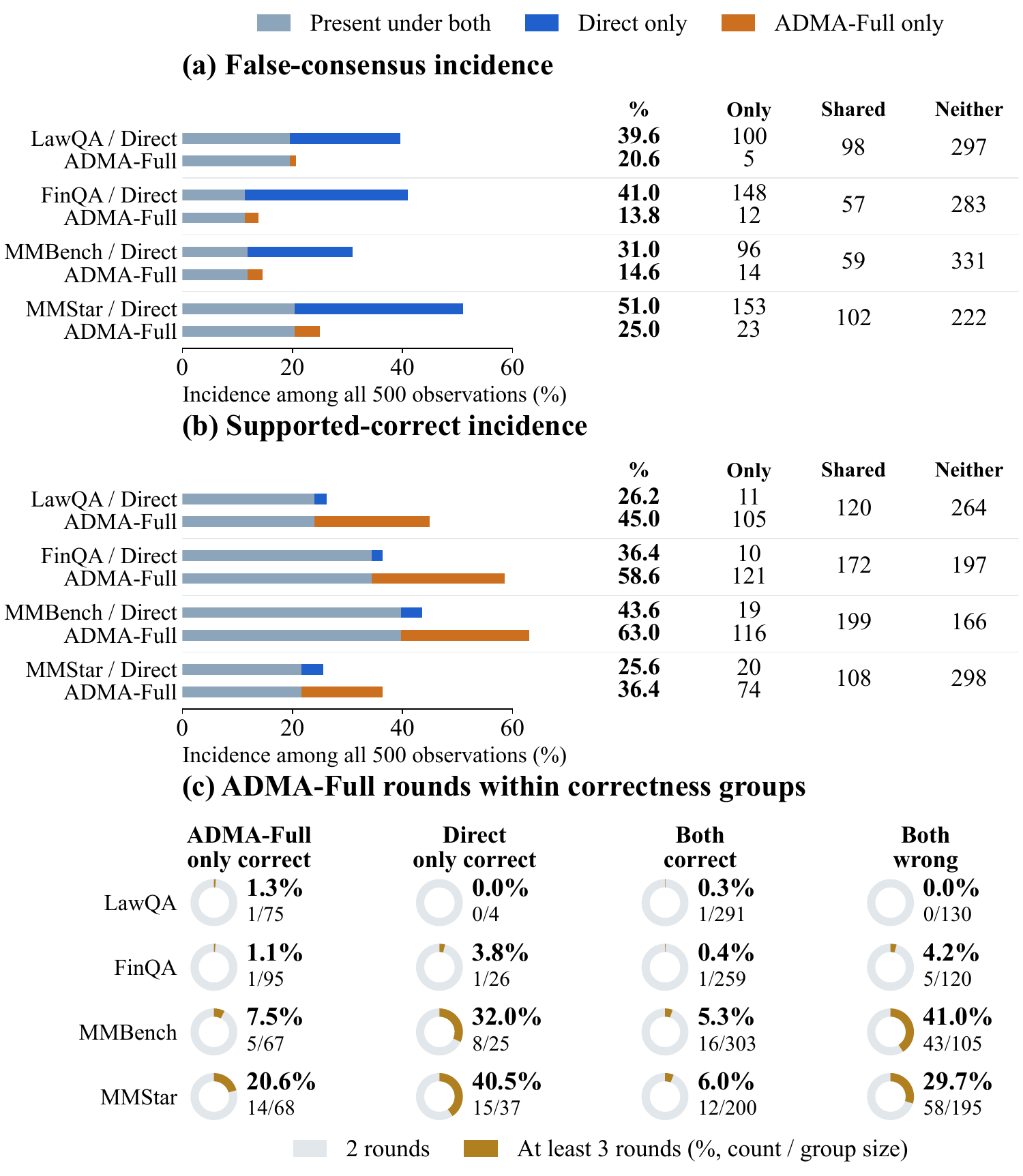}
\caption{Reliability and deliberation depth within Qwen3.5. \textbf{a}--\textbf{b}, paired false-consensus and supported-correct incidence among all 500 observations per task, separated into shared and method-exclusive outcomes. \textbf{c}, ADMA-Full round distributions within paired correctness groups defined against Direct. Gray denotes two rounds and gold at least three; labels report the latter's percentage, count, and group denominator. MMBench uses the maximum across permutations, and failure-affected traces are excluded.}
\label{fig:reliability}
\end{figure}

Table~\ref{tab:raw_strength_results} gives FCR and ESR within their conditional denominators. Figure~\ref{fig:reliability}(a,b) instead shows shared and method-exclusive events over the full panel, revealing which predictions produce those rate changes.

Deliberation depth also differs by task and outcome. Two rounds account for 99.6\% of LawQA and 98.4\% of FinQA traces, compared with 85.6\% on MMBench and 80.2\% on MMStar. Most text questions therefore stop after the required revision, whereas visual tasks continue more often. Continuation concentrates among jointly missed MMBench questions and also among recovered MMStar questions. Extra rounds consequently mark both successful correction and unresolved difficulty.

The Qwen3.6 visual results further separate correctness from support. Accuracy increases on both tasks, while ESR changes from 95.7\% to 95.5\% on MMBench and from 90.6\% to 87.8\% on MMStar. More correct answers can therefore coexist with a lower supported proportion within the enlarged correct subset.

\subsection{Resource--accuracy trade-off}

Finally, we quantify the computation associated with adaptive deliberation. Median paired ratios relative to Direct range from 14.3 to 18.7 for tokens and from 7.8 to 16.5 for latency. Their distributions show substantial question-level variation. Qwen3.5 MMStar, for example, has a lower median token ratio than LawQA but a wider 10th--90th percentile span.

\begin{figure}[!t]
\centering
\includegraphics[width=\textwidth]{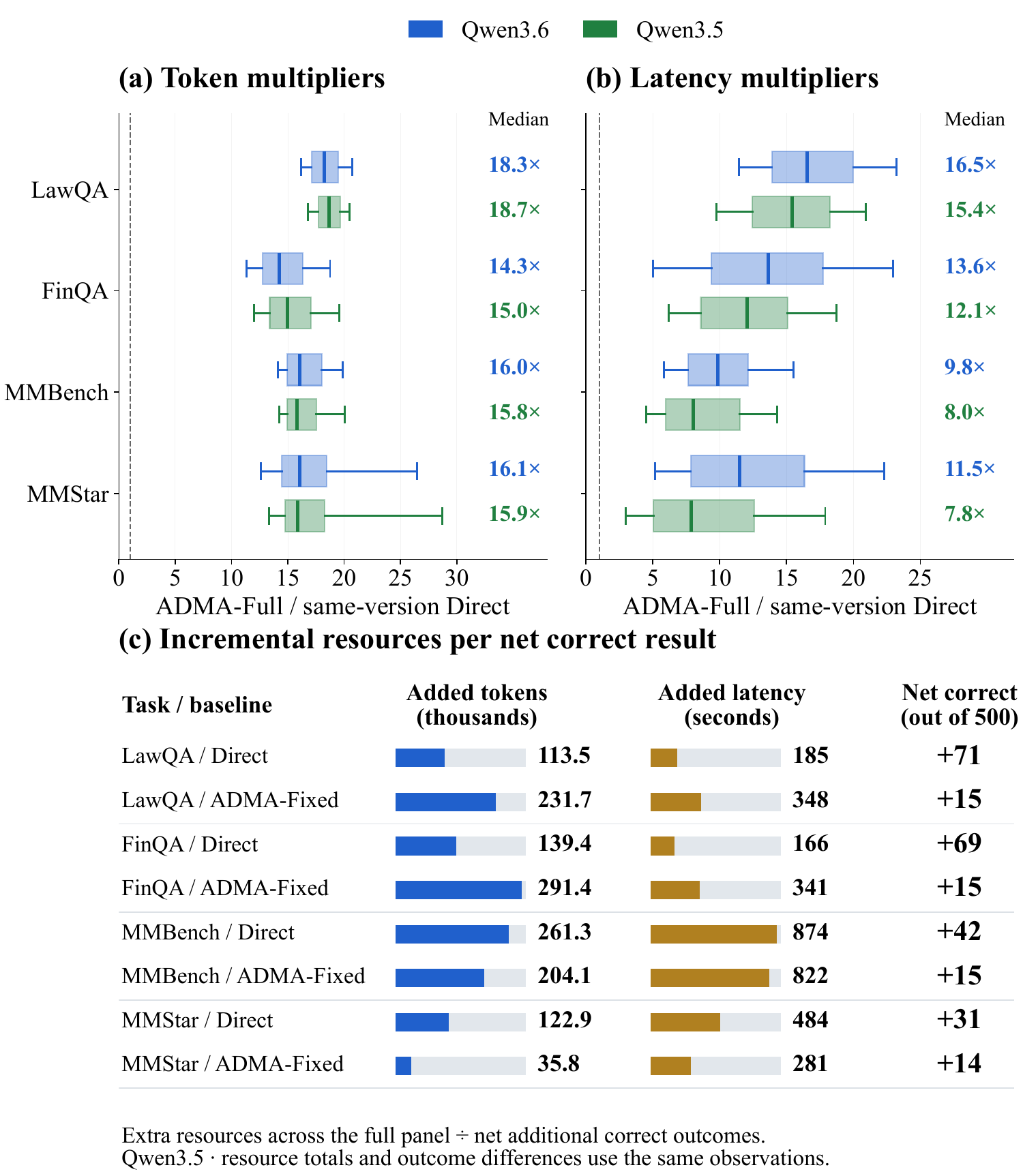}
\caption{Resource demand of ADMA-Full. \textbf{a}--\textbf{b}, within-version token and latency ratios to Direct over failure-free pairs. Boxes span the 25th--75th percentiles, whiskers the 10th--90th percentiles, central lines mark medians, and dashed lines denote parity. \textbf{c}, total Qwen3.5 resource increments over the stated baseline divided by the net additional correct count among 500 paired observations. MMBench resources are summed over permutations, and latency is the sum of elapsed inference times.}
\label{fig:cost}
\end{figure}

On FinQA, each net additional correct outcome over Direct corresponds to approximately 139,400 additional tokens and 166 seconds of summed prediction time. Relative to Fixed, the smaller net gain raises these values to approximately 291,400 tokens and 341 seconds. These ratios include resources spent on recovered, unchanged, and regressed outcomes, so they describe protocol-level expenditure rather than correction-only cost.

\section{Discussion}
\label{sec:discussion}

The results distinguish a recognition model's capability from the policy that organizes its use. With Qwen3.5 fixed, ADMA-Full improves every task over Direct and ADMA-Fixed. Because the same controller is transferred without policy updates, these gains arise from changed interaction rather than task-model training. Adaptive deliberation can therefore operate as a reusable inference component within the tested model family and task set.

The policy is most useful when the task exposes checkable alternatives. Multi-step and mixed-source FinQA questions contain intermediate operations and distinct information sources, consistent with their larger gains. Visual-category differences reinforce this interpretation, because mathematics improves while coarse perception declines on MMStar. Deliberation appears better suited to revisable reasoning than to errors dominated by perceptual recognition.

Cross-version results define a second boundary. Qwen3.5 ADMA-Full surpasses Qwen3.6 Direct on FinQA and MMStar, matches it on MMBench, and remains lower on LawQA. Interaction can offset a model-version gap when correction opportunities exceed the initial capability difference. It cannot be treated as a general substitute for a stronger Direct model.

Reliability results clarify why agreement alone is insufficient. Under Qwen3.5, fewer false-consensus events and more supported correct outcomes accompany the accuracy gains. Additional rounds, however, occur among both recovered and unresolved cases. Neither consensus nor duration independently establishes decision quality; both require outcome-aware assessment of the trace.

The gains carry substantial computational demand. Median token and latency ratios remain far above Direct, and expenditure per net gain depends on the comparator and outcome balance. These results favor selective deployment where the expected correction value exceeds the resource cost.

The evidence is bounded by one selected controller, two inference repeats, fixed 250-question panels, and model-based process assessment. Paired resampling quantifies panel sensitivity, while subgroup and cost-per-gain analyses remain descriptive. The conclusions therefore concern the specified models, protocols, assessors, and task panels. Within this scope, the joint accuracy, reliability, and cost analyses support adaptive control while identifying the conditions that govern its value.

\section{Conclusion}
\label{sec:conclusion}

ADMA learns how frozen language-model roles should propose, challenge, arbitrate, and coordinate as a recognition trace develops. Centralized training captures interdependent role effects, while decentralized actors apply role-specific controls without updating the task model.

With Qwen3.5 fixed, ADMA-Full improves four benchmarks by 6.2--14.2 percentage points over Direct and 2.8--3.0 points over ADMA-Fixed. Paired outcomes attribute the net improvement to recoveries exceeding regressions. Process assessment further records fewer false-consensus events and more supported correct outcomes. Transfer to Qwen3.6 and comparison with its Direct predictions show that these benefits remain task dependent.

The study therefore supports adaptive deliberation as a controllable inference policy, not as a universal replacement for stronger models or cheaper direct inference. Its practical value depends on the task's correction opportunities and the acceptable resource budget. More selective control and expert-calibrated process assessment are the next requirements for extending this approach beyond the evaluated setting.

\section*{Declaration of generative AI and AI-assisted technologies in the manuscript preparation process}

During the preparation of this work, the authors used OpenAI Codex to improve language clarity and readability. After using this tool, the authors reviewed and edited the manuscript as needed and take full responsibility for its content.

\section*{CRediT authorship contribution statement}

\textbf{Yuan Liu}: Conceptualization, Methodology, Software, Writing -- original draft; \textbf{Haoyue Peng}: Methodology, Investigation, Formal analysis, Visualization; \textbf{Fanliang Hu}: Data curation, Validation, Formal analysis; \textbf{Lei Huang}: Resources, Validation, Writing -- review \& editing; \textbf{Wenli Cai}: Conceptualization, Supervision, Project administration, Writing -- review \& editing.

\section*{Data availability}

The public benchmarks are available from their original repositories. Code, controller weights, and derived evaluation records are available in the \href{https://github.com/Alex-liuyuan/ADMA-Self-Learning-Deliberation-Control/tree/d21f85c24c90c352ef78a23b934acdcdc18bde1d}{project repository}. Detailed analyses are provided in the Supplementary Material; corresponding numerical records and editable table sources are supplied in the Supplementary Data.

\section*{Declaration of competing interest}

The authors declare that they have no known competing financial interests or personal relationships that could have appeared to influence the work reported in this paper.

\section*{Acknowledgement}

This work was supported by the Chongqing ``Yuyue Action'' Top Talent Special Project (Grant No. YYXM002), hosted by Chongqing University Cancer Hospital.

\bibliographystyle{elsarticle-num}
\bibliography{references}

\clearpage
\appendix
\setcounter{table}{0}
\renewcommand{\thetable}{S\arabic{table}}
\renewcommand{\thesection}{S\arabic{section}}
\setlength{\tabcolsep}{4pt}
\renewcommand{\arraystretch}{1.2}
This material provides repeat-level results, paired outcome decompositions, task-stratified analyses, and resource statistics for the experiments reported in the manuscript. The analyses use the same archived runs as the main results. The accompanying numerical records enable the reported counts, rates, and resource summaries to be checked without reproducing benchmark questions or images.

\section{Evaluation and controller specification}

The four evaluation panels contain 250 base questions each. Each model--protocol combination is evaluated twice on the same panel. Thus, a task contributes 500 question--repeat observations, while paired uncertainty analysis retains 250 base-question clusters. LawQA denotes CaseHOLD accessed through LexGLUE. MMBench uses strict CircularEval, requiring all available option permutations to be correct. FinQA and MMStar retain their task-specific answer evaluation.

\begin{table}[H]
\centering
\caption{Controller and evaluation settings used in the manuscript.}
\label{tab:settings}
\begin{tabular}{@{}p{4.7cm}p{10.3cm}@{}}
\toprule
Component & Setting \\
\midrule
Evaluation task models & Qwen3.5-35B-A3B and Qwen3.6-35B-A3B \\
Frozen training task model & Gemini-3.1-Flash-Lite-Preview \\
Role inventory & Proposer, challenger, arbiter, coordinator \\
Observation per role & 22 features, comprising 16 shared and six role-specific features \\
Action dimensions & 4, 4, 4, and 5, respectively, in the role order above \\
Deliberation horizon & At most six rounds; Full requires at least two, Fixed permits one \\
Training panel & 250 LawQA and 250 FinQA training questions; 500 episodes \\
Development and selection & 50 questions per task; balanced 20-question checkpoint subset, evaluated every 100 episodes \\
Selected checkpoint & Episode 400, by development accuracy with quality resolving ties \\
Learning coefficients & Discount 0.95; Polyak update 0.01; minibatch 256 \\
Transfer & One frozen controller, without policy updates for either Qwen version or either visual task \\
\bottomrule
\end{tabular}
\end{table}

Direct uses one response per ordinary question or CircularEval permutation. ADMA-Fixed uses constant role controls and is evaluated only with Qwen3.5. ADMA-Full applies the learned controls and challenger-guided revision. Full--Fixed therefore compares complete protocols, including the different minimum revision schedules. In the following tables, Qwen3.5 and Qwen3.6 abbreviate the full model identifiers in Table~\ref{tab:settings}; Fixed and Full abbreviate ADMA-Fixed and ADMA-Full.

\clearpage
\section{Repeat-level recognition and process assessment}

Tables~\ref{tab:runs_weak} and \ref{tab:runs_strong} disaggregate the mean and standard deviation reported in the main manuscript. Repeats use the same selected controller and question identifiers. They characterize inference variability rather than independent controller training runs. Every accuracy denominator remains 250, including incomplete predictions.

For each repeat, let $C$ be the number of correct predictions, $F$ the number of false-consensus events, and $S$ the number of supported correct predictions. The reported percentages are
\begin{equation}
\mathrm{Accuracy}=100\frac{C}{250},\qquad
\mathrm{FCR}=100\frac{F}{250-C},\qquad
\mathrm{ESR}=100\frac{S}{C}.
\end{equation}
FCR and ESR therefore have different, outcome-dependent denominators. Only completed predictions with completed assessments register process events; unavailable assessments are retained as missing.

\begin{table}[H]
\centering
\caption{Repeat-level performance with Qwen3.5-35B-A3B.}
\label{tab:runs_weak}
\begin{tabular}{@{}llrrrrr@{}}
\toprule
Task & Protocol & Repeat & Correct & Accuracy (\%) & FCR (\%) & ESR (\%) \\
\midrule
LawQA & Direct & 1 & 145 & 58.0 & 96.2 & 43.4 \\
LawQA & Direct & 2 & 150 & 60.0 & 97.0 & 45.3 \\
LawQA & ADMA-Fixed & 1 & 173 & 69.2 & 84.4 & 53.2 \\
LawQA & ADMA-Fixed & 2 & 178 & 71.2 & 88.9 & 55.1 \\
LawQA & ADMA-Full & 1 & 182 & 72.8 & 73.5 & 60.4 \\
LawQA & ADMA-Full & 2 & 184 & 73.6 & 80.3 & 62.5 \\
FinQA & Direct & 1 & 140 & 56.0 & 96.4 & 62.9 \\
FinQA & Direct & 2 & 145 & 58.0 & 94.3 & 64.8 \\
FinQA & ADMA-Fixed & 1 & 169 & 67.6 & 61.7 & 78.1 \\
FinQA & ADMA-Fixed & 2 & 170 & 68.0 & 65.0 & 79.4 \\
FinQA & ADMA-Full & 1 & 176 & 70.4 & 47.3 & 82.4 \\
FinQA & ADMA-Full & 2 & 178 & 71.2 & 47.2 & 83.1 \\
MMBench & Direct & 1 & 162 & 64.8 & 90.9 & 66.7 \\
MMBench & Direct & 2 & 166 & 66.4 & 89.3 & 66.3 \\
MMBench & ADMA-Fixed & 1 & 178 & 71.2 & 58.3 & 78.7 \\
MMBench & ADMA-Fixed & 2 & 177 & 70.8 & 61.6 & 80.2 \\
MMBench & ADMA-Full & 1 & 184 & 73.6 & 57.6 & 84.2 \\
MMBench & ADMA-Full & 2 & 186 & 74.4 & 54.7 & 86.0 \\
MMStar & Direct & 1 & 117 & 46.8 & 97.7 & 53.0 \\
MMStar & Direct & 2 & 120 & 48.0 & 96.2 & 55.0 \\
MMStar & ADMA-Fixed & 1 & 127 & 50.8 & 66.7 & 57.5 \\
MMStar & ADMA-Fixed & 2 & 127 & 50.8 & 69.9 & 59.8 \\
MMStar & ADMA-Full & 1 & 132 & 52.8 & 53.4 & 66.7 \\
MMStar & ADMA-Full & 2 & 136 & 54.4 & 54.4 & 69.1 \\
\bottomrule
\end{tabular}
\par\smallskip\begin{minipage}{\textwidth}\footnotesize Each row contains 250 base questions. FCR is conditional on incorrect predictions; ESR is conditional on correct predictions. -- denotes an unavailable assessment.\end{minipage}
\end{table}

The Qwen3.5 process assessor is Gemini-3.1-Flash-Lite-Preview. The assessor receives the input, answer options, method-neutral trace, and image when applicable, but not the reference answer. Adequate support requires relevant observations, reasoning, or verifiable calculations. Independent support requires at least two substantively different, independently checkable reasoning paths.

\clearpage
\begin{table}[H]
\centering
\caption{Repeat-level performance with Qwen3.6-35B-A3B.}
\label{tab:runs_strong}
\begin{tabular}{@{}llrrrrr@{}}
\toprule
Task & Protocol & Repeat & Correct & Accuracy (\%) & FCR (\%) & ESR (\%) \\
\midrule
LawQA & Direct & 1 & 194 & 77.6 & -- & -- \\
LawQA & Direct & 2 & 192 & 76.8 & -- & -- \\
LawQA & ADMA-Full & 1 & 189 & 75.6 & -- & -- \\
LawQA & ADMA-Full & 2 & 185 & 74.0 & -- & -- \\
FinQA & Direct & 1 & 146 & 58.4 & 72.1 & 96.6 \\
FinQA & Direct & 2 & 141 & 56.4 & 88.1 & 95.7 \\
FinQA & ADMA-Full & 1 & 181 & 72.4 & 68.1 & 97.8 \\
FinQA & ADMA-Full & 2 & 178 & 71.2 & 59.7 & 98.9 \\
MMBench & Direct & 1 & 183 & 73.2 & 82.1 & 96.2 \\
MMBench & Direct & 2 & 187 & 74.8 & 85.7 & 95.2 \\
MMBench & ADMA-Full & 1 & 191 & 76.4 & 49.2 & 95.8 \\
MMBench & ADMA-Full & 2 & 190 & 76.0 & 36.7 & 95.3 \\
MMStar & Direct & 1 & 129 & 51.6 & 93.4 & 92.2 \\
MMStar & Direct & 2 & 118 & 47.2 & 92.4 & 89.0 \\
MMStar & ADMA-Full & 1 & 149 & 59.6 & 56.4 & 86.6 \\
MMStar & ADMA-Full & 2 & 145 & 58.0 & 53.3 & 89.0 \\
\bottomrule
\end{tabular}
\par\smallskip\begin{minipage}{\textwidth}\footnotesize Each row contains 250 base questions. FCR is conditional on incorrect predictions; ESR is conditional on correct predictions. -- denotes an unavailable assessment.\end{minipage}
\end{table}

The Qwen3.6 process assessor is Gemini-3.1-Flash-Lite. Its LawQA assessments are unavailable and are not imputed. Conditional reliability comparisons are made within a task-model version under its designated assessor.

\begin{table}[H]
\centering
\caption{Incomplete predictions with Qwen3.6-35B-A3B.}
\label{tab:failures}
\begin{tabular}{@{}llrr@{}}
\toprule
Task & Protocol & Repeat 1 & Repeat 2 \\
\midrule
LawQA & Direct & 1 & 1 \\
LawQA & ADMA-Full & 1 & 1 \\
FinQA & Direct & 0 & 0 \\
FinQA & ADMA-Full & 0 & 0 \\
MMBench & Direct & 4 & 4 \\
MMBench & ADMA-Full & 5 & 4 \\
MMStar & Direct & 3 & 3 \\
MMStar & ADMA-Full & 4 & 5 \\
\bottomrule
\end{tabular}
\par\smallskip\begin{minipage}{\textwidth}\footnotesize Counts use 250 questions per repeat. A failed required CircularEval permutation makes its base question incomplete. No incomplete predictions occur in the archived Qwen3.5 runs.\end{minipage}
\end{table}

An incomplete prediction counts as incorrect in primary accuracy. For process assessment, an event must additionally satisfy the completion criteria. Concentrated commitment follows the manuscript definition, using high final-answer confidence at a threshold of 0.80, consistent semantic answers across CircularEval permutations, or explicitly reported consensus. Under Direct inference, this identifies the single-response analogue of the diagnostic.

\clearpage
\section{Paired effect estimates and panel sensitivity}

For question $i$ and repeat $r$, let $a_{ir}$ and $b_{ir}$ denote binary correctness for the compared protocol and its reference. The cluster effect and overall difference in percentage points (pp) are
\begin{equation}
d_i=\frac{1}{2}\sum_{r=1}^{2}(a_{ir}-b_{ir}),\qquad
\widehat{\Delta}=\frac{100}{250}\sum_{i=1}^{250}d_i.
\end{equation}
The paired bootstrap samples 250 base-question clusters with replacement, keeping the two repeats together. The 2.5th and 97.5th percentiles of 10,000 resampled differences form the reported 95\% ranges. These ranges characterize sensitivity to panel composition, conditional on the evaluated controller and protocols. They are reported separately for each contrast without multiplicity adjustment.

\begin{table}[H]
\centering
\caption{Within-version accuracy differences for ADMA-Full minus the indicated baseline.}
\label{tab:paired}
\begin{tabular}{@{}lllrrrr@{}}
\toprule
Version & Task & Baseline & Repeat 1 & Repeat 2 & Mean & 95\% range \\
\midrule
Qwen3.5 & LawQA & Direct & +14.8 & +13.6 & +14.2 & [10.0, 18.6] \\
Qwen3.5 & FinQA & Direct & +14.4 & +13.2 & +13.8 & [8.4, 19.2] \\
Qwen3.5 & MMBench & Direct & +8.8 & +8.0 & +8.4 & [3.6, 13.4] \\
Qwen3.5 & MMStar & Direct & +6.0 & +6.4 & +6.2 & [1.2, 11.2] \\
Qwen3.5 & LawQA & Fixed & +3.6 & +2.4 & +3.0 & [0.4, 5.8] \\
Qwen3.5 & FinQA & Fixed & +2.8 & +3.2 & +3.0 & [0.2, 6.0] \\
Qwen3.5 & MMBench & Fixed & +2.4 & +3.6 & +3.0 & [0.8, 5.4] \\
Qwen3.5 & MMStar & Fixed & +2.0 & +3.6 & +2.8 & [-1.4, 7.0] \\
Qwen3.6 & LawQA & Direct & -2.0 & -2.8 & -2.4 & [-6.0, 1.4] \\
Qwen3.6 & FinQA & Direct & +14.0 & +14.8 & +14.4 & [9.2, 19.6] \\
Qwen3.6 & MMBench & Direct & +3.2 & +1.2 & +2.2 & [-1.6, 6.0] \\
Qwen3.6 & MMStar & Direct & +8.0 & +10.8 & +9.4 & [4.4, 14.4] \\
\bottomrule
\end{tabular}
\end{table}

\begin{table}[H]
\centering
\caption{Cross-version accuracy differences for Qwen3.5 ADMA-Full minus Qwen3.6 Direct.}
\label{tab:cross}
\begin{tabular}{@{}lrrrr@{}}
\toprule
Task & Repeat 1 & Repeat 2 & Mean & 95\% range \\
\midrule
LawQA & -4.8 & -3.2 & -4.0 & [-8.0, -0.2] \\
FinQA & +12.0 & +14.8 & +13.4 & [8.2, 18.6] \\
MMBench & +0.4 & -0.4 & 0.0 & [-4.6, 5.0] \\
MMStar & +1.2 & +7.2 & +4.2 & [-1.6, 10.0] \\
\bottomrule
\end{tabular}
\end{table}

The within-Qwen3.5 Full--Fixed ranges are positive for LawQA, FinQA, and MMBench. The cross-version FinQA margin has a positive range, whereas the positive MMStar mean has a range spanning zero. The cross-version analysis consequently describes task-dependent compensation; it does not establish uniform dominance over the later model. MMBench and MMStar are reported separately because their source collections may overlap.

\clearpage
\section{Error recovery and consistency across repeats}

Table~\ref{tab:transitions} separates newly correct outcomes from lost correct outcomes, preserving all matched observations. The accuracy difference equals the net gained count divided by 500. Table~\ref{tab:consistency} returns to the 250 base questions to distinguish gains repeated twice from gains appearing in only one repeat.

\begin{table}[H]
\centering
\caption{Paired correctness for Qwen3.5 ADMA-Full relative to Qwen3.6 Direct.}
\label{tab:transitions}
\begin{tabular}{@{}lrrrrr@{}}
\toprule
Task & Both correct & Recovered & Lost & Both incorrect & Net gain \\
\midrule
LawQA & 344 & 22 & 42 & 92 & -20 \\
FinQA & 262 & 92 & 25 & 121 & 67 \\
MMBench & 320 & 50 & 50 & 80 & 0 \\
MMStar & 177 & 91 & 70 & 162 & 21 \\
\bottomrule
\end{tabular}
\par\smallskip\begin{minipage}{\textwidth}\footnotesize Each task has 500 paired question--repeat observations. Recovered denotes a Direct error corrected by Full; lost denotes a Direct success missed by Full.\end{minipage}
\end{table}

\begin{table}[H]
\centering
\caption{Repeat consistency of the cross-version paired difference.}
\label{tab:consistency}
\begin{tabular}{@{}lrrrrrr@{}}
\toprule
Task & Win twice & Win once & Reversal & Lose once & Lose twice & No change \\
\midrule
LawQA & 4 & 14 & 0 & 12 & 15 & 205 \\
FinQA & 31 & 29 & 1 & 10 & 7 & 172 \\
MMBench & 15 & 17 & 3 & 23 & 12 & 180 \\
MMStar & 21 & 39 & 10 & 28 & 16 & 136 \\
\bottomrule
\end{tabular}
\par\smallskip\begin{minipage}{\textwidth}\footnotesize Counts sum to 250 base questions. Win/lose once means a signed difference in one repeat and a tie in the other. Reversal means opposite signed differences across repeats. No change means a tie in each repeat, whether both methods are correct or both are incorrect.\end{minipage}
\end{table}

\begin{table}[H]
\centering
\caption{Joint correctness across the three Qwen3.5 protocols.}
\label{tab:paths}
\begin{tabular}{@{}lllrrrr@{}}
\toprule
Direct & Fixed & Full & LawQA & FinQA & MMBench & MMStar \\
\midrule
Incorrect & Incorrect & Incorrect & 120 & 111 & 101 & 167 \\
Incorrect & Incorrect & Correct & 24 & 21 & 14 & 39 \\
Incorrect & Correct & Incorrect & 10 & 9 & 4 & 28 \\
Incorrect & Correct & Correct & 51 & 74 & 53 & 29 \\
Correct & Incorrect & Incorrect & 4 & 21 & 20 & 23 \\
Correct & Incorrect & Correct & 1 & 8 & 10 & 17 \\
Correct & Correct & Incorrect & 0 & 5 & 5 & 14 \\
Correct & Correct & Correct & 290 & 251 & 293 & 183 \\
\bottomrule
\end{tabular}
\par\smallskip\begin{minipage}{\textwidth}\footnotesize Each task column sums to 500. Rows describe separate protocol evaluations of matched questions, not sequential rounds of one deliberation.\end{minipage}
\end{table}

Table~\ref{tab:paths} supplies the eight joint outcome patterns behind the main manuscript's protocol-flow analysis. It permits reconstruction of any pairwise Qwen3.5 correctness comparison while retaining the common evaluation panel. These counts describe associations between separate inference protocols and do not attribute an individual correction to a particular learned action.

\clearpage
\section{Task-stratified paired outcomes}

The FinQA partitions in Table~\ref{tab:subgroups_finqa} are defined separately by the first annotated operation family, the annotated information source, and the number of reasoning steps. Addition includes summation. Each partition contains all 250 FinQA questions once; counts must not be summed across different partitions. Gains and losses pool the two repeats. The visual categories in Table~\ref{tab:subgroups_vision} use benchmark annotations and likewise partition each panel once.

\begin{table}[H]
\centering
\caption{Qwen3.5 ADMA-Full versus Direct by FinQA stratum.}
\label{tab:subgroups_finqa}
\begin{tabular}{@{}lp{5.1cm}rrrrr@{}}
\toprule
Partition & Group & $n$ & Gained & Lost & Tied & Gain (pp) \\
\midrule
Operation & Addition & 37 & 11 & 7 & 56 & +5.4 \\
Operation & Division & 72 & 17 & 8 & 119 & +6.2 \\
Operation & Other operations & 26 & 8 & 2 & 42 & +11.5 \\
Operation & Subtraction & 115 & 59 & 9 & 162 & +21.7 \\
Source & Mixed & 37 & 13 & 1 & 60 & +16.2 \\
Source & Table & 151 & 61 & 18 & 223 & +14.2 \\
Source & Text & 62 & 21 & 7 & 96 & +11.3 \\
Steps & Multiple steps & 116 & 53 & 12 & 167 & +17.7 \\
Steps & One step & 134 & 42 & 14 & 212 & +10.4 \\
\bottomrule
\end{tabular}
\par\smallskip\begin{minipage}{\textwidth}\footnotesize $n$ counts base questions. Gained, lost, and tied count the $2n$ paired observations across repeats. Tied combines both-correct and both-incorrect outcomes.\end{minipage}
\end{table}

\begin{table}[H]
\centering
\caption{Qwen3.5 ADMA-Full versus Direct by visual benchmark category.}
\label{tab:subgroups_vision}
\begin{tabular}{@{}lp{5.1cm}rrrrr@{}}
\toprule
Task & Group & $n$ & Gained & Lost & Tied & Gain (pp) \\
\midrule
MMBench & attribute reasoning & 43 & 6 & 6 & 74 & 0.0 \\
MMBench & coarse perception & 63 & 16 & 5 & 105 & +8.7 \\
MMBench & Fine-grained perception (cross) & 31 & 8 & 4 & 50 & +6.5 \\
MMBench & Fine-grained perception (instance) & 63 & 20 & 3 & 103 & +13.5 \\
MMBench & logic reasoning & 25 & 8 & 4 & 38 & +8.0 \\
MMBench & relation reasoning & 25 & 9 & 3 & 38 & +12.0 \\
MMStar & coarse perception & 42 & 3 & 11 & 70 & -9.5 \\
MMStar & fine-grained perception & 42 & 14 & 4 & 66 & +11.9 \\
MMStar & instance reasoning & 41 & 9 & 7 & 66 & +2.4 \\
MMStar & logical reasoning & 42 & 9 & 5 & 70 & +4.8 \\
MMStar & math & 41 & 17 & 4 & 61 & +15.9 \\
MMStar & science \& technology & 42 & 16 & 6 & 62 & +11.9 \\
\bottomrule
\end{tabular}
\par\smallskip\begin{minipage}{\textwidth}\footnotesize $n$ counts base questions. Gained, lost, and tied count the $2n$ paired observations across repeats. Tied combines both-correct and both-incorrect outcomes.\end{minipage}
\end{table}

These descriptive decompositions show where recovery outweighs regression within the evaluated panels. They support interpretation of the aggregate results without treating small subgroups as independently powered confirmatory experiments.

\clearpage
\section{Reliability events and deliberation depth}

Table~\ref{tab:reliability} uses the full 500-observation task denominator, complementing the conditional FCR and ESR in Section~S2. An event may occur under neither method, Direct alone, Full alone, or both methods. This decomposition distinguishes an actual change in event count from a change driven by the size of the correct or incorrect subset.

\begin{table}[H]
\centering
\caption{Paired process-assessment events under Qwen3.5.}
\label{tab:reliability}
\begin{tabular}{@{}llrrrrr@{}}
\toprule
Task & Event & Neither & Direct only & Full only & Both & Net \\
\midrule
LawQA & False consensus & 297 & 100 & 5 & 98 & -95 \\
LawQA & Supported correct & 264 & 11 & 105 & 120 & 94 \\
FinQA & False consensus & 283 & 148 & 12 & 57 & -136 \\
FinQA & Supported correct & 197 & 10 & 121 & 172 & 111 \\
MMBench & False consensus & 331 & 96 & 14 & 59 & -82 \\
MMBench & Supported correct & 166 & 19 & 116 & 199 & 97 \\
MMStar & False consensus & 222 & 153 & 23 & 102 & -130 \\
MMStar & Supported correct & 298 & 20 & 74 & 108 & 54 \\
\bottomrule
\end{tabular}
\par\smallskip\begin{minipage}{\textwidth}\footnotesize Each row sums to 500 across the four event states. Net is Full minus Direct event count; a negative net favors Full for false consensus and a positive net favors Full for supported correctness.\end{minipage}
\end{table}

\begin{table}[H]
\centering
\caption{ADMA-Full depth within Qwen3.5 correctness groups defined against Direct.}
\label{tab:rounds}
\begin{tabular}{@{}llrrrr@{}}
\toprule
Task & Outcome & $N$ & Two rounds & $\geq 3$ rounds & $\geq 3$ (\%) \\
\midrule
LawQA & Both correct & 291 & 290 & 1 & 0.3 \\
LawQA & Recovered & 75 & 74 & 1 & 1.3 \\
LawQA & Lost & 4 & 4 & 0 & 0.0 \\
LawQA & Both incorrect & 130 & 130 & 0 & 0.0 \\
FinQA & Both correct & 259 & 258 & 1 & 0.4 \\
FinQA & Recovered & 95 & 94 & 1 & 1.1 \\
FinQA & Lost & 26 & 25 & 1 & 3.8 \\
FinQA & Both incorrect & 120 & 115 & 5 & 4.2 \\
MMBench & Both correct & 303 & 287 & 16 & 5.3 \\
MMBench & Recovered & 67 & 62 & 5 & 7.5 \\
MMBench & Lost & 25 & 17 & 8 & 32.0 \\
MMBench & Both incorrect & 105 & 62 & 43 & 41.0 \\
MMStar & Both correct & 200 & 188 & 12 & 6.0 \\
MMStar & Recovered & 68 & 54 & 14 & 20.6 \\
MMStar & Lost & 37 & 22 & 15 & 40.5 \\
MMStar & Both incorrect & 195 & 137 & 58 & 29.7 \\
\bottomrule
\end{tabular}
\par\smallskip\begin{minipage}{\textwidth}\footnotesize $N$ counts complete question--repeat traces in each group. CircularEval depth is the maximum over its permutations. Percentages use the group-specific denominator.\end{minipage}
\end{table}

Round allocation is an observed property of the completed traces. A longer trace can accompany either a recovered outcome or an unresolved error, so the association between duration and correctness is not interpreted as a causal effect of additional rounds.

\clearpage
\section{Resource distributions and incremental expenditure}

For paired observation $j$, the resource ratio is $R_j=U_j^{\mathrm{Full}}/U_j^{\mathrm{Direct}}$, where $U$ is token use or elapsed inference time. Table~\ref{tab:resources} pools ratios across repeats after excluding pairs with an incomplete prediction. CircularEval token use and elapsed time are summed over permutations. Its deliberation depth, used in Section~S6, is instead the maximum round count over permutations.

\begin{table}[H]
\centering
\caption{Within-version ADMA-Full-to-Direct resource ratios.}
\label{tab:resources}
\begin{tabular}{@{}lllrrrrrr@{}}
\toprule
Version & Task & Resource & $N$ & $P_{10}$ & $P_{25}$ & $P_{50}$ & $P_{75}$ & $P_{90}$ \\
\midrule
Qwen3.5 & LawQA & Tokens & 500 & 16.8 & 17.7 & 18.7 & 19.6 & 20.5 \\
Qwen3.5 & LawQA & Latency & 500 & 9.7 & 12.5 & 15.4 & 18.3 & 20.9 \\
Qwen3.5 & FinQA & Tokens & 500 & 12.0 & 13.4 & 15.0 & 17.0 & 19.6 \\
Qwen3.5 & FinQA & Latency & 500 & 6.2 & 8.6 & 12.1 & 15.1 & 18.8 \\
Qwen3.5 & MMBench & Tokens & 500 & 14.2 & 14.9 & 15.8 & 17.5 & 20.0 \\
Qwen3.5 & MMBench & Latency & 500 & 4.5 & 5.9 & 8.0 & 11.5 & 14.3 \\
Qwen3.5 & MMStar & Tokens & 500 & 13.3 & 14.7 & 15.9 & 18.3 & 28.7 \\
Qwen3.5 & MMStar & Latency & 500 & 3.0 & 5.0 & 7.8 & 12.6 & 17.9 \\
Qwen3.6 & LawQA & Tokens & 498 & 16.2 & 17.1 & 18.3 & 19.5 & 20.7 \\
Qwen3.6 & LawQA & Latency & 498 & 11.5 & 13.9 & 16.5 & 20.0 & 23.2 \\
Qwen3.6 & FinQA & Tokens & 500 & 11.3 & 12.7 & 14.3 & 16.3 & 18.7 \\
Qwen3.6 & FinQA & Latency & 500 & 5.0 & 9.4 & 13.6 & 17.7 & 23.0 \\
Qwen3.6 & MMBench & Tokens & 491 & 14.1 & 14.9 & 16.0 & 18.0 & 19.9 \\
Qwen3.6 & MMBench & Latency & 491 & 5.9 & 7.6 & 9.8 & 12.1 & 15.5 \\
Qwen3.6 & MMStar & Tokens & 491 & 12.6 & 14.5 & 16.1 & 18.4 & 26.5 \\
Qwen3.6 & MMStar & Latency & 491 & 5.2 & 7.8 & 11.5 & 16.3 & 22.3 \\
\bottomrule
\end{tabular}
\par\smallskip\begin{minipage}{\textwidth}\footnotesize Ratios are dimensionless. $N$ counts paired complete observations across both repeats. Each percentile is computed over per-observation ratios, not from a ratio of aggregate means.\end{minipage}
\end{table}

Percentiles use the nearest indexed observation at position $\operatorname{round}((N-1)p)$ in the zero-indexed ordered sample, with ties rounded to the even integer. This convention matches the archived analysis. Latency represents observed elapsed inference time, including the evaluated execution conditions; it is not a hardware-independent estimate of model computation.

\begin{table}[H]
\centering
\caption{Cross-version comparison restricted to matched complete predictions.}
\label{tab:sensitivity}
\begin{tabular}{@{}lrrrr@{}}
\toprule
Task & Paired observations & Qwen3.6 Direct (\%) & Qwen3.5 Full (\%) & Margin (pp) \\
\midrule
LawQA & 498 & 77.5 & 73.1 & -4.4 \\
FinQA & 500 & 57.4 & 70.8 & +13.4 \\
MMBench & 492 & 75.2 & 74.4 & -0.8 \\
MMStar & 494 & 50.0 & 54.3 & +4.3 \\
\bottomrule
\end{tabular}
\par\smallskip\begin{minipage}{\textwidth}\footnotesize Both methods are evaluated on exactly the same retained observations. This sensitivity analysis does not replace the fixed-denominator primary analysis.\end{minipage}
\end{table}

The matched complete-prediction analysis removes the same observations from both sides of each cross-version comparison. It separates the effect of incomplete predictions from the primary fixed-denominator result. The complete-pair resource distributions and this accuracy sensitivity analysis serve different purposes and use the appropriate pairing for each comparison.

\clearpage
\begin{table}[H]
\centering
\caption{Qwen3.5 resource increments per net additional correct outcome.}
\label{tab:incremental}
\begin{tabular}{@{}llrrrrr@{}}
\toprule
Task & Baseline & $\Delta C$ & $\Delta T$ (M) & $\Delta L$ (ks) & Tokens/$\Delta C$ (k) & Seconds/$\Delta C$ \\
\midrule
LawQA & Direct & 71 & 8.059 & 13.15 & 113.5 & 185.2 \\
LawQA & Fixed & 15 & 3.476 & 5.21 & 231.7 & 347.6 \\
FinQA & Direct & 69 & 9.616 & 11.43 & 139.4 & 165.6 \\
FinQA & Fixed & 15 & 4.371 & 5.12 & 291.4 & 341.4 \\
MMBench & Direct & 42 & 10.976 & 36.72 & 261.3 & 874.2 \\
MMBench & Fixed & 15 & 3.062 & 12.34 & 204.1 & 822.5 \\
MMStar & Direct & 31 & 3.810 & 14.99 & 122.9 & 483.5 \\
MMStar & Fixed & 14 & 0.501 & 3.93 & 35.8 & 280.9 \\
\bottomrule
\end{tabular}
\par\smallskip\begin{minipage}{\textwidth}\footnotesize All 500 observations per task enter each total. $\Delta C$ is the net correct-count increment, $\Delta T$ the token increment, and $\Delta L$ the summed latency increment. M, k, and ks denote millions, thousands, and thousands of seconds.\end{minipage}
\end{table}

For a specified baseline $B$, expenditure per net additional correct outcome is
\begin{equation}
K_U=\frac{\sum_{j=1}^{500}(U_j^{\mathrm{Full}}-U_j^{B})}
{\sum_{j=1}^{500}(a_j^{\mathrm{Full}}-a_j^{B})}.
\end{equation}
The numerator includes every observation, including unchanged and regressed outcomes. This measure is therefore a panel-level incremental expenditure, not the average cost of a successfully corrected question. The denominator is positive for all eight Qwen3.5 contrasts in Table~\ref{tab:incremental}. Summed elapsed times quantify total measured prediction time and should not be interpreted as the duration of a parallel evaluation job.

\section{Data provenance and numerical reconstruction}

The accompanying data archive contains 40 run-level records and 10,000 base-question--protocol--repeat records across the evaluated combinations. Records preserve task identifiers, model versions, protocol labels, correctness, process-assessment completion, and resource measurements. No benchmark question text, images, or private service configuration is included.

Each result table is generated from the bundled data rather than entered manually; the settings in Table~\ref{tab:settings} follow the main manuscript. Automated checks reconcile run-level accuracy and conditional rates with question-level events, verify paired recovery and loss counts, confirm partition totals, and reconstruct reliability transitions, round allocations, resource quantiles, and incremental expenditure. A provenance manifest records the source-file and bundled-data checksums. These checks establish consistency with the archived evaluation records.

The source package provides the editable document and a standard-library table generator. The data dictionary defines the machine-readable fields, model aliases, units, missing values, and links between files and tables. The qualitative MMStar illustration in the main manuscript concerns a separate archived Gemini run and is not included in these aggregate Qwen statistics.

\end{document}